\documentclass[10pt,twocolumn,letterpaper]{article}

\usepackage{cvpr}
\usepackage{times}
\usepackage{epsfig}
\usepackage{graphicx}
\usepackage{amsmath}
\usepackage{amssymb}
\usepackage{multirow}
\usepackage{booktabs}
\usepackage{bm}
\usepackage{authblk}

\usepackage[breaklinks=true,bookmarks=false]{hyperref}

\cvprfinalcopy 

\def\cvprPaperID{****} 

\begin{document}

\title{IoU-uniform R-CNN: Breaking Through the Limitations of RPN}

\author[1]{Li Zhu}
\author[1]{Zihao Xie}
\author[4]{Liman Liu}
\author[3]{Bo Tao}
\author[1,2]{Wenbing Tao \thanks{Corresponding author}}

\affil[1]{National Key Laboratory of Science and Technology on Multi-spectral Information
	Processing, School of Artifcial Intelligence and Automation, Huazhong University of
	Science and Technology}
\affil[2]{Shenzhen Huazhong University of Science and Technology Research Institute, Shenzhen, 518057, China}
\affil[3]{State Key Laboratory of Digital Manufacturing Equipment and Technology, School of Mechanical Science and Engineering, Huazhong University of Science and Technology, Wuhan, Hubei 430074, PR China}
\affil[4]{School of Biomedical Engineering, South-Central University for Nationalities, Wuhan 430074, China \authorcr \tt\small \{lizhu2016,zihaoxie,taobo,wenbingtao\}@hust.edu.cn limanliu@mail.scuec.edu.cn}

\maketitle

\begin{abstract}
Region Proposal Network (RPN) is the cornerstone of two-stage object detectors, it generates a sparse set of object proposals and alleviates the extrem foreground-background class imbalance problem during training. However, we find that the potential of the detector has not been fully exploited due to the IoU distribution imbalance and inadequate quantity of the training samples generated by RPN. With the increasing intersection over union (IoU), the exponentially smaller numbers of positive samples would lead to the distribution skewed towards lower IoUs, which hinders the optimization of detector at high IoU levels. In this paper, to break through the limitations of RPN, we propose IoU-Uniform R-CNN, a simple but effective method that directly generates training samples with uniform IoU distribution for the regression branch as well as the IoU prediction branch. Besides, we improve the performance of IoU prediction branch by eliminating the feature offsets of RoIs at inference, which helps the NMS procedure by preserving accurately localized bounding box. Extensive experiments on the PASCAL VOC and MS COCO dataset show the effectiveness of our method, as well as its compatibility and adaptivity to many object detection architectures. The code is made publicly available at \url{https://github.com/zl1994/IoU-Uniform-R-CNN}.
\end{abstract}

\section{Introduction}
Recent years has witnessed the remarkable progress in object detection thanks to the advance of the deep convolution networks \cite{alexnet, VGG, inception,resnet}. Among them, the two-stage approach is the leading paradigm in the deep learning era of object detection and Region Proposal Network (RPN)\cite{faster_r-cnn} is the cornerstone of two-stage object detectors. It generates region proposals from a dense set of anchors and these proposals are further refined by subsequent region-wise R-CNN subnetwork. Specially, \cite{focalloss} pointed out that the RPN alleviates the extrem foreground-background class imbalance problem by filtering out the majority of negative locations and this is the central cause that the performance of two-stage detectors is better than one-stage detectors. Although RPN plays a important role in the existing model structure, our study reveals that the potential of the two-stage detectors has not been fully exploited due to the limitations of RPN. 

\begin{figure}[t]
	\begin{center}
		\includegraphics[width=1.0\linewidth]{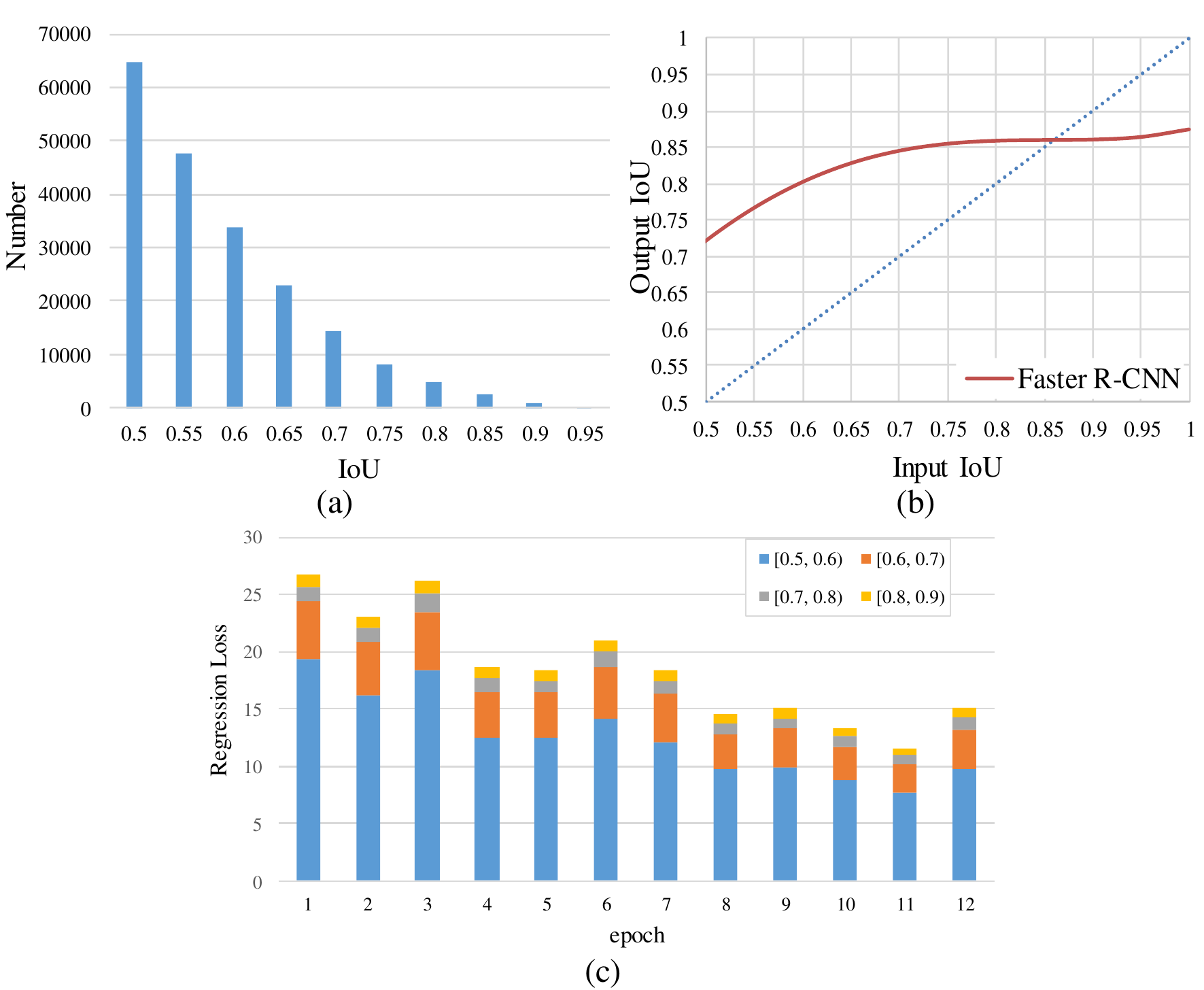}
	\end{center}
	\caption{(a) IoU histogram of the RoIs generated by the RPN. (b) Localization performance of object detectors. (c) Composition of regression loss.}
	\label{fig:long}
	\label{fig:onecol}
\end{figure}

Figure 1 (a) shows the IoU histogram of the Region of interests (RoIs) generated by the RPN. With the increase of IoU, the number of RoIs decreases sharply, leading to the IoU distribution imbalance. As the subsequent R-CNN takes the RoIs as training samples, the distribution of training samples is naturally skewed towards lower IoUs. What's more, the total number of positive samples per image is no more than 100 during the training procedure while the total number of training samples is 512. We argue that the IoU distribution imbalance and inadequate quantity of the positive samples hinder the optimization of detector, especially at high IoU levels. It can be seen in Figure 1 (b), we plot the IoU of the RoIs with their corresponding GT bounding boxes before and after regression. The localization accuracy gains of RoIs, after the refinement of regressor, are mainly concentrated at low IoU levels and it even decays at high IoU levels. We attribute this to the loss imbalance during training. Figure 1 (c) illustrates the composition of the regression loss during training. It can be seen that the low IoU RoIs comprise the majority of the loss and dominate the gradients. As a result, the detector optimized at low IoU level is not necessarily optimal at other level, which influence the overall performance of the detector. 

In order to solve the above problems, Cascade R-CNN \cite{cascade} proposed a multi-stage object detection framework. The detectors are trained stage by stage and the training samples of following stages are the output of previous stage, as the output IoU of a regressor is almost invariably better than the input IoU, the detector can obtain enough samples at different IoU levels, which improve the overall performance of the detector. Although Cascade R-CNN obtained solid improvements, the multi-stage way is not flexible enough. Libra R-CNN \cite{librarcnn} proposed IoU-balanced sampling, which is a simple but effective method, to alleviate the distribution imbalance among hard negative samples. However, because of the exponentially vanishing high IoU samples, it is hard to get uniform IoU distribution for positive samples by this sampling strategy.

There comes a question: For a two-stage detector, do the training samples of R-CNN must come from the output of RPN ? The answer is No. In this paper, instead of taking positive samples from RPN, we propose to add controllable jitter to each GT bounding box to directly generate positive training samples for R-CNN. So that we can simply and effectively obtain adequate uniform distributed training samples for not only the regression branch, but also the IoU prediction branch \cite{iou-net}. Our experiments have shown that the uniform IoU distribution, formed by generated samples, can greatly promote the performance of the regression and IoU prediction branch.

\begin{figure}[t]
	\begin{center}
		\includegraphics[width=1.0\linewidth]{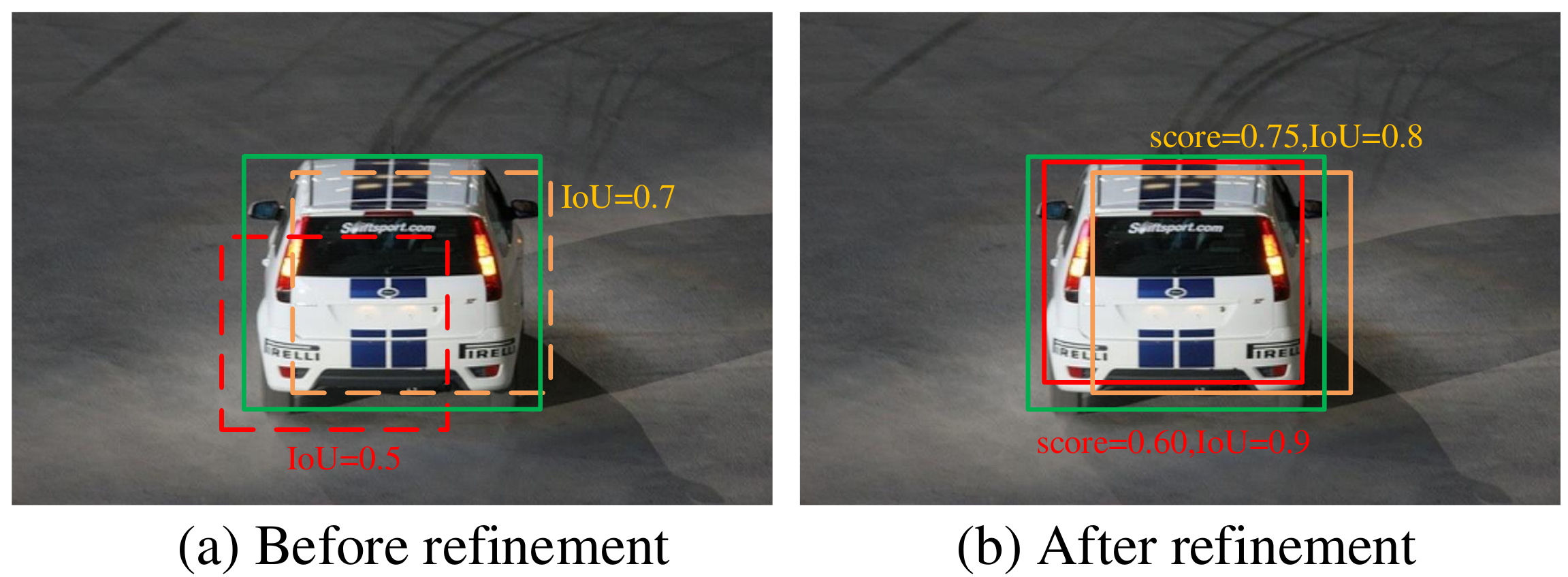}
	\end{center}
	\caption{Illustration of misalignment between predicted IoU and localization accuracy. (a) Show two proposals that covering the same GT bbox. The IoU of the red one is smaller than the orange one (b) The red bbox come from behind after the refinement of second stage. But the score of the red one is still lower than the orange one, leading to the more accurate localized bbox suppressed during NMS procedure.}
	\label{fig:long}
	\label{fig:onecol}
\end{figure}

What's more, as non-maximum suppression (NMS) procedure is a critical post-processing procedure to filter redundant bounding boxes, IoU-Net \cite{iou-net} pointed out that the misalignment between classification confidence and localization accuracy may lead to accurately localized bounding boxes being suppressed by less accurate ones in the NMS. To solve this problem, IoU-Net proposed a IoU-prediction branch to predict the IoU between the predicted bounding box and the corresponding ground truth bounding box. The predicted IoU replaces the classification score as the metric for ranking the bounding boxes. Nevertheless, we argue that there is still a mismatch in IoU-prediction branch. During training, the input of the IoU predictor is the RoI feature at the current position, and the IoU predictor outputs the predicted IoU of the RoI with its corresponding GT bounding box. But when it comes to the test phase, the predicted IoU is assigned to the bounding box, which has been moved to a new position after the refinement of RoI by the regression branch. The shift of RoI position also brings feature offset. It is the feature offset of RoI that result in a misalignment between predicted IoU and localization accuracy. It is shown in Figure 2, the red and orange dotted box are the proposals and they cover the same GT bbox. Although the IoU of the red box is smaller than the orange box, the red box come from behind after the refinement of R-CNN network. However, as the input of following branch is still the feature at the position before the regression, the red box score is still lower than the orange box score, which would lead to the more accurate one be suppressed during NMS. In our paper, we further improves the performance of IoU prediction branch by eliminating the feature offsets of RoIs at inference without training. Experiments have shown it can boost the performance of detector and the stronger the IoU prediction branch is, the more gains it brings.

Our main contributions are summarized as follows: (1) Our study reveals the importance of solving the limitations of RPN and our proposed IoU-uniform R-CNN can alleviate the IoU distribution imbalance and inadequate training samples by generating samples with uniform IoU distribution. (2) We improve the performance of IoU prediction branch by eliminating the feature offsets of RoIs at inference. (3) Our proposed method consistently obtains significant improvements over multiple state-of-the-art detectors. Specially, without bells and whistles, it achieves 2.4 AP improvement than Faster R-CNN (with ResNet-101-FPN backbone) on MS COCO dataset. 

\section{Related Work}
\textbf{Development of the model architecture.} Nowadays, with the deep learning techniques have been widely applied to various computer vision tasks, convolution neural networks (CNNs) based approaches have prevailed on object detection task and the model architectures are also constantly evolving. The CNN based detectors were first introduced by R-CNN \cite{rcnn}. Its derivatives, Fast R-CNN \cite{fast_r-cnn} and Faster R-CNN \cite{faster_r-cnn}, further improve the speed and performance by introducing the RoI Pooling and Region Proposal Network (RPN)  module. The method described above and the following R-FCN \cite{rfcn} and Cascade R-CNN \cite{cascade} can be classified as two-stage methods. They first obtain a sparse set of proposals and then classify and refine these proposals at the second stage. On the other hand, single-stage detectors are popularized by YOLO \cite{yolo,yolo9000} and SSD \cite{ssd}. They focus on high efficiency and treat object detection as a single shot problem. The performance gap between two-stage methods and one-stage methods was narrowed by RetinaNet \cite{focalloss}. Besides, with the help of pyramid structure of FPN \cite{FPN}, the detectors enhance its  feature extraction capabilities, further improving performance. Recently, as the single-stage and two-stage detection frameworks becomes mature, the anchor-free methods have become a new research hotspot. Instead of using anchor boxes, they predict bounding boxes in a per-pixel prediction fashion \cite{fcos,FSAF} or keypoint-based fashion \cite{cornernet,centernet}.

\textbf{Imbalance problems in object detection.} Nowadays, as the model architecture becomes mature, more and more research has resort to improve the training process of the detector. Under such circumstance, the sample imbalance problems during training have attracted growing attentions. \cite{oksuz2019imbalance} review the deep-learning-era object detection literature and identify 8 different imbalance problems. Numerous studies have also shown that mitigating sample imbalance, especially the Forground-Background imbalance, would bring significant gains to the performance of the detector. For example, in order to alleviate the Forground-Background imbalance, Focal loss \cite{focalloss}, Gradient Harmonizing Mechanism (GHM) \cite{GHM} solves it by a soft sampling way, which suppresses the gradient originating from easy positives and negatives. And SSD \cite{ssd}, OHEM \cite{OHEM} restrict the imbalance by hard example mining. However, the IoU distribution imbalance has received relatively less attention in object detection. Cascade R-CNN \cite{cascade} tried to solve this problem by a cascade framework. RoIs were iteratively refined and the detector can obtain enough sample at different IoU levels ultimately. Libra R-CNN \cite{librarcnn} proposed IoU-balanced sampling to alleviate the IoU distribution imbalance among hard negative samples.

\textbf{Improvement of Duplicate Removal.} Duplicate Removal is an essential postprocessing procedure of object detectors for removing duplicated bounding boxes. Its efficacy heavily affects the final performance. The most widely used algorithm is non-maximum suppression (NMS). It iteratively selects proposals according to the confidence score (usually the classification score) and suppresses overlapped proposals. However, the classification score is not accurate enough to guarantee preserving the most accurate detection results. Instead of directly eliminating overlapped proposals, Soft-NMS \cite{softnms} decays the bounding box scores and Softer NMS \cite{softer} averages the selected boxes in a softer way. Fitness NMS \cite{fitnessnms} introduces the localization information while ranking the bounding box into ranking confidence. Different from existing NMS algorithms, Prime sample \cite{primesample} investigates the sample importance and make the classifier more prone to give high scores to high IoU proposals. IoU-Net \cite{iou-net} claims that it is not proper to use classification scores as the ranking criterion, it proposed a IoU-prediction branch to predict the IoU between the predicted bounding box and the corresponding ground truth. The predicted IoU replaces the classification score as the metric for ranking the bounding boxes.

\section{Methodology}
In this section, we will illustrate the proposed IoU-uniform R-CNN for object detection. As our goal is to break through the limitations of RPN during the training of detectors, we first replace the RoI training samples with generated samples to obtain more powerful regressor and IoU predictor. Furthermore we simply tune loss weight of different IoU intervals to control the balance of regression loss composition. Then we propose to eliminate the feature offsets of RoI during the inference of IoU-prediction branch. With more powerful regressor and IoU predictor, IoU-uniform R-CNN achieves superior performance. All components will be elaborated below.
\subsection{Generate positive samples with uniform IoU distribution}
We first revisit the pipeline of two-stage approach. As illustrated in Figure 4 (a), the Region Proposal Networks (RPN)
generates a sparse set of proposals that should cover all forground objects while filtering out the majority of negative locations. Then at the second stage, a region-wise subnetwork is designed to refine these proposals by further classification and regression. The whole network is trained end-to-end and the region-wise subnetwork takes the output of RPN as training samples. The region-wise subnetwork is expected to do it well among different quality proposals but things go athwart. Figure 3 shows the average localization improvement of proposals from different IoU intervals after refinement. We can find that, with the increase of IoU, the gain of refinement get smaller and the performance even get worse at high IoU levels. As discussed in section 1, the performance imbalance may come from the loss imbalance during training. As most of the train samples are from low IoU levels (IoU\textless  0.7), the low IoU RoIs comprise the majority of the loss and dominate the gradients.

\begin{figure}[t]
	\begin{center}
		\includegraphics[width=0.8\linewidth]{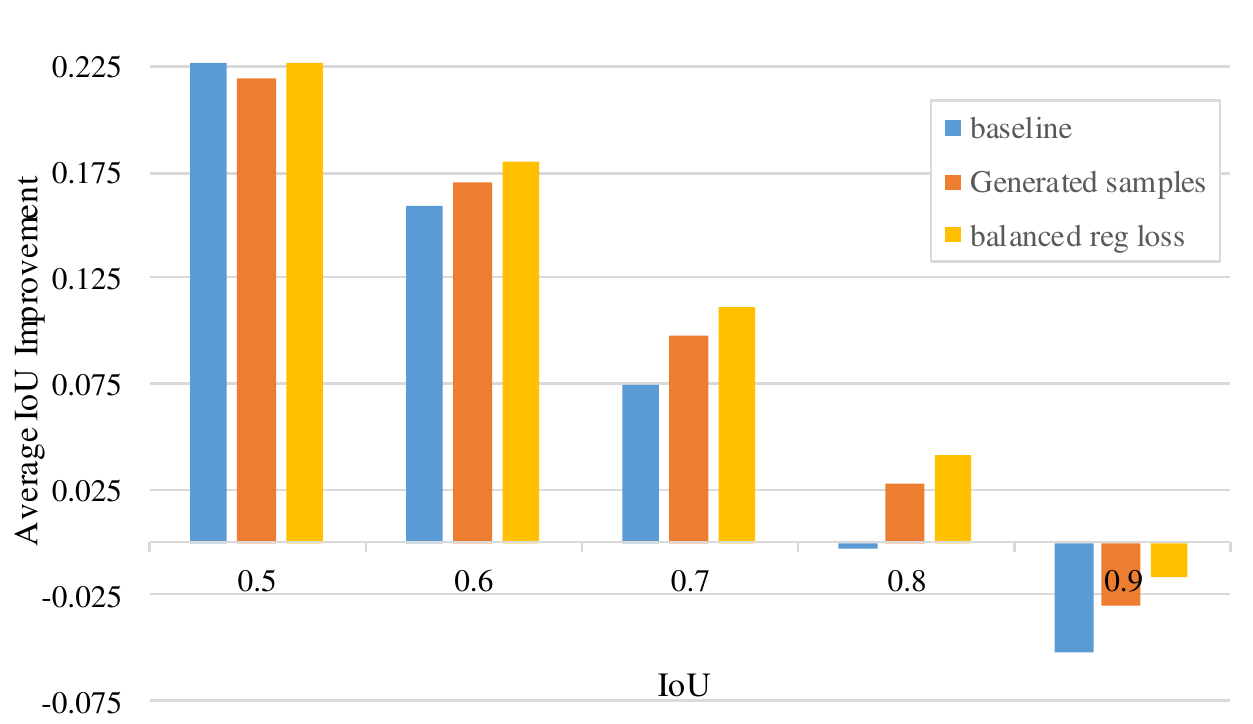}
	\end{center}
	\caption{The average localization improvement of proposals from different IoU intervals after refinement.}
	\label{fig:long}
	\label{fig:onecol}
\end{figure}

A natural solution for alleviating the imbalance is to resample or tune the loss weight of different IoU intervals. However, as the quantity gap between low IoU level and high IoU level is too wide and the number of positive samples is not enough (no more than 100 positive samples per image), it is hard to get uniform IoU distribution for positive samples by resampling strategy. And it is also hard to determine appropriate weights to balance the composition of regression loss. 

In this paper, in order to obtain samples with uniform IoU distribution for region-wise subnetwork, we propose to directly generate training samples around each GT bounding box, instead of taking proposals from RPN. We first divide the IoU into $\bm{N}$ intervals and then we generate samples for each bounding box at each interval by adding controllable jitters. Given an image with $\bm{K}$ annotated ground-truths, a bounding box is represented by $\boldsymbol{b}=(b_{x},b_{y},b_{w},b_{h},b_{x})$. A generated sample is determined by:
\begin{equation}\label{eq:mixture}
\begin{array}{l}
b{'_x} = {b_x} + {b_w} * random\_offset\_x\\
b{'_y} = {b_y} + {b_h} * random\_offset\_y\\
b{'_w} = {b_w} * random\_w\\
b{'_h} = {b_h} * random\_h
\end{array}
\end{equation}

\begin{figure}[t]
	\begin{center}
		\includegraphics[width=0.8\linewidth]{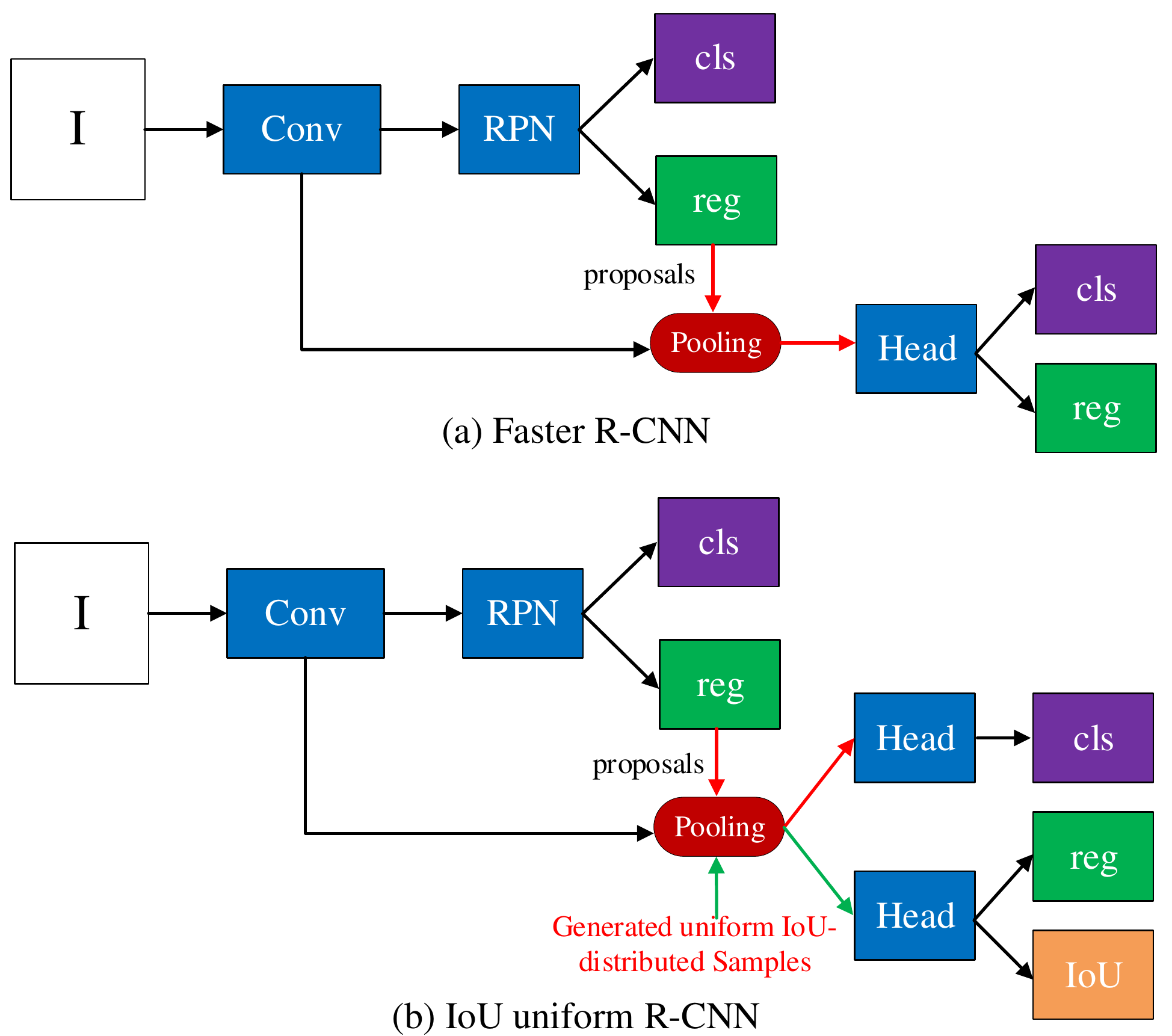}
	\end{center}
	\caption{The architectures of different frameworks.}
	\label{fig:long}
	\label{fig:onecol}
\end{figure}

In order to obtain enough samples from limitted number of attempts, the random range among intervals depends on the IoU level of generated RoI samples. Precisely, for obtaining higher IoU samples, the random range should get smaller. Besides, to guarantee the validity of generated samples, we only keep the RoIs which has the maximum IoU over the current GT bounding box. As we keep $\bm{M}$ samples for each IoU interval per GT bouding box, the overall number of the training samples for one image is $\bm{K*N*M}$ and we obtain a totally uniform IoU distribution at last. To this end, our training pipeline is shown in Figure 4 (b), the generated IoU uniformly distributed samples are used to train not only the regression branch but also the IoU-prediction branch. The following experiments will show it can greatly promote the performance of both the regression and IoU prediction branch. As for the classification branch, it still take the output of RPN as training samples, for the big difference of RoI distribution between training and test stage is harmful to the performance of classifiers.

Although the number of samples for among different IoU intervals is now the same, we can still find the regression loss imbalance. It may caused by the initialization. As the layers are randomly initialized with normal distributions, where the mean is set to 0 and standard deviation 0.001. The size of the output value of regression branch is concentrated around 0 at the beginning. For the regression is trained to predict the offset between RoI and GT bounding box, This initialization will lead to the the low IoU samples dominate total loss in early training because the offset of low IoU RoI is natural bigger then high IoU ones. As the imbalance problem has been largely mitigated and we already have enough training samples for all IoU intervals, tuning the regression weights according to the IoU of proposals have becoming feasible and easy. The weighted regression loss is shown in Equ. 2, $\bm{L}$ is the commonly used smooth L1 loss. The main idea of the reweighting strategy is to down-weight training samples with low IoU and up-weight samples with high IoU. With the further ease of the imbalance problem, we obtain more robust model for accurate localization.
\begin{equation}
{L_{reg}} = \sum\limits_j^N {\sum\limits_i^M {{w_j} * L } (} {d_i},{\hat d_i})
\end{equation}

\subsection{Eliminating the feature offsets of RoIs}
The IoU-prediction branch \cite{iou-net} was proposed to predict the localization confidence for each detected bounding box and it is more sensitive to localization accuracy. Thus, the feature offsets of RoIs can not be neglected. It is shown in Figure 5, to eliminate the feature offsets, we set the output bounding box of the region-wise subnetwork as the new RoI and obtain the new RoI features by RoIAlign pooling \cite{mask_r-cnn}. Then the new RoI feaures was sent to the region-wise subnetwork again to obtain the ultimate IoU prediction result. It is the twice feature extraction of RoIs at inference that help us to eliminate the feature offsets of RoIs without training.

Ideally, we expect the IoU score of bbox candidates to replace the classification score as the suppression criterion of NMS algorithm, but the IoU score of numerous background bounding box is not credible enough during test, because IoU-prediction branch is only trained by positive samples whose IoU is above 0.5. However, the messy situation of background samples would bring much trouble to the training procedure of IoU-prediction branch and influence the accuracy on those high IoU bounding boxes which play a key role in driving the detection performance. Therefore, it is inappropriate to train the IoU-prediction branch by both positive and negative samples. To remedy this, we set the multiplication of the IoU score and classification score as the final score of the suppression criterion of the NMS algorithm, for the classification branch can help to suppress the background bboxes with low classification score.

\begin{figure*}[t]
	\begin{center}
		\includegraphics[width=1.0\linewidth]{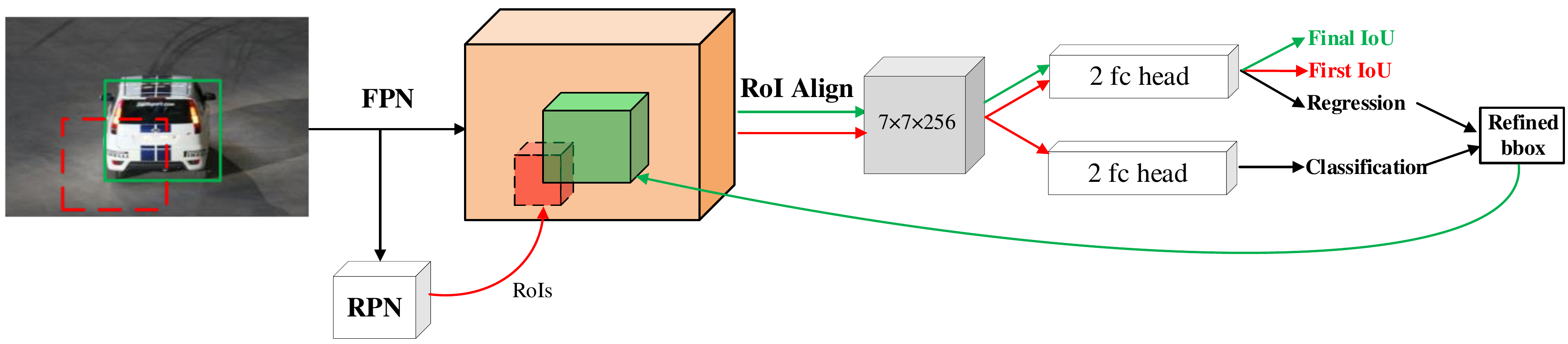}
	\end{center}
	\caption{Illustration of eliminating the feature offsets of RoIs. The red lines represent the first time RoI feature extraction and IoU prediction. The green lines represent the second time RoI feature extraction and IoU prediction.}
	\label{fig:long}
	\label{fig:onecol}
\end{figure*}

\begin{table*}[htbp]
	\centering
	\caption{Main results on PASCAL VOC 2007 test set}
		\begin{tabular}{ccccccccc}
			\toprule
			Backbone & Detector & Our method & AP & AP$_{50}$ & AP$_{60}$  & AP$_{70}$  & AP$_{80}$  & AP$_{90}$ \\
			\midrule
			\multirow{4}[2]{*}{ResNet-50-FPN} & Faster R-CNN & No & 50.2 & 79.1 & 75.6 & 63.4 & 42.4 & 10.8 \\
			& Faster R-CNN & Yes   & 55.4 & 79.8 & 75.9 & 67.2 & 51.6 & 23.0 \\
			& Cascade R-CNN & No    & 54.9 & 79.2 & 74.3 & 66.2 & 51.7 & 23.2 \\
			& Cascade R-CNN & Yes   & 56.1 & 77.8 & 74.1 & 66.1 & 53.4 & 29.4 \\
			\midrule
			\multirow{4}[2]{*}{ResNet-101-FPN} & Faster R-CNN & No & 52.8& 82.2 & 77.1 & 66.3 & 45.9 & 12.1 \\
			& Faster R-CNN & Yes    & 57.6 & 81.4 & 77.6 & 69.2 & 54.1 & 25.4 \\
			& Cascade R-CNN & No    & 57.7 & 81.9 & 77.5 & 68.7 & 56.3 & 25.9\\
			& Cascade R-CNN & Yes   & 57.6 & 78.4 & 74.4 & 67.2 & 55.0 & 31.7  \\
			\bottomrule
	\end{tabular}%
	\label{tab:addlabel}%
\end{table*}%

\section{Experiments}
We comprehensively evaluate our method on two widely used benchmarks for the object detection task: MS COCO \cite{coco} and PASCAL VOC \cite{voc}. In particular, MS COCO is a large scale dataset with 80 object categories. It consists of 115k images for training (train-2017), 5k images for validation (val-2017), 20k for testing without provided annotations. We use the train split for training and report the performance on validation and test-dev split. PASCAL VOC is another dataset for evaluating our method. We use the union of VOC2007 and VOC2012 trainval as training set, which contains 16551 images and objects from 20 pre-defined categories annotated with bounding boxes. We evaluate our models on the VOC 2007 test set. 

\begin{table*}[htbp]
	\centering
	\caption{Main results  on COCO validation set}
		\begin{tabular}{ccccccccc}
			\toprule
			Backbone & Detector & Our method & AP & AP$_{50}$ & AP$_{75}$  & AP$_{S}$ & AP$_{M}$ & AP$_{L}$ \\
			\midrule
			\multirow{4}[2]{*}{ResNet-50-FPN} & Faster R-CNN & No & 36.4 & 58.4 & 39.1 & 21.5 & 40.0 & 46.6 \\
			& Faster R-CNN & Yes   & 39.1 & 57.7 & 42.2 & 22.5 & 42.6 & 50.7 \\
			& Cascade R-CNN & No    & 40.4 & 58.5 & 43.9 & 21.5 & 43.7 & 53.8 \\
			& Cascade R-CNN & Yes   & 40.9 & 58.1 & 43.9 & 23.1 & 44.1 & 53.7 \\
			\midrule
			\multirow{4}[2]{*}{ResNet-101-FPN} & Faster R-CNN & No & 38.5 & 60.3 & 41.6 & 22.3 & 43.0 & 49.8 \\
			& Faster R-CNN & Yes   & 40.9 & 59.7  & 43.8  & 22.9  & 44.9  & 54.4 \\
			& Cascade R-CNN & No   & 42.0 & 60.3 & 45.9 & 23.2 & 45.9 & 23.2 \\
			& Cascade R-CNN & Yes  & 42.4 & 59.7 & 45.7 & 23.8 & 45.9 & 56.5  \\
			\bottomrule
	\end{tabular}%
	\label{tab:addlabel}%
\end{table*}%

\begin{table*}[htbp]
	\centering
	\caption{Comparison with state-of-the-art detectors on COCO test-dev.}
		\begin{tabular}{lccccccc}
			\toprule
			Method & backbone & AP & AP$_{50}$ & AP$_{75}$  & AP$_{S}$ & AP$_{M}$ & AP$_{L}$ \\
			\midrule
			YOLOv2 & DarkNet-19 & 21.6  & 44.0  & 19.2  & 5.0   & 22.4  & 35.5  \\
			YOLOv3 & DarkNet-53 & 33.0  & 57.9  & 34.4  & 18.3  & 35.4  & 41.9  \\
			SSD513 & ResNet-101 & 31.2  & 50.4  & 33.3  & 10.2  & 34.5  & 49.8  \\
			RetinaNet & ResNet-101 & 39.1  & 59.1  & 42.3  & 21.8  & 42.7  & 50.2  \\
			\midrule
			Faster R-CNN & ResNet-101-FPN & 38.8 & 60.9 & 42.3 & 22.3 & 42.2 & 48.6  \\
			Faster R-CNN by G-RMI \cite{G-RMI} & Inception-ResNet-v2 & 34.7  & 55.5  & 36.7  & 13.5  & 38.1  & 52.0  \\
			Faster R-CNN w/TDM \cite{TDM} & Inception-ResNet-v2-TDM & 36.8  & 57.5  & 39.2  & 16.2  & 39.8  & 52.1  \\
			Deformable R-FCN \cite{deformable} & Aligned-Inception-ResNet & 37.5  & 58.0  & 40.8  & 19.4  & 40.1  & 52.5  \\
			Mask R-CNN \cite{mask_r-cnn} & ResNet-101-FPN & 38.2  & 60.3  & 41.7  & 20.1  & 41.1  & 50.2  \\
			Cascade R-CNN & ResNet-50-FPN & 40.7 & 59.3 & 44.1 & 23.1 & 43.6 & 51.4  \\
			Cascade R-CNN & ResNet-101-FPN & 42.4  & 61.1  & 46.1  & 23.6  & 45.4  & 54.1  \\
			Libra R-CNN & ResNet-101-FPN & 40.3  & 61.3  & 43.9  & 22.9  & 43.1  & 51.0  \\
			IoU-Net & ResNet-101-FPN & 40.6  & 59.0  & -  & -  & -  & -  \\
			\midrule
			Faster R-CNN+IoU-uniform R-CNN & ResNet-50-FPN & 39.0  & 57.8  & 42.0  & 22.4  & 41.9  & 48.7  \\
			Faster R-CNN+IoU-uniform R-CNN & ResNet-101-FPN & 41.2  & 60.1 & 44.3  & 23.6  & 44.1  & 52.2  \\
			Cascade R-CNN+IoU-uniform R-CNN & ResNet-50-FPN & 41.3 & 58.8 & 44.4 & 23.8 & 43.9 & 52.0  \\
			Cascade R-CNN+IoU-uniform R-CNN & ResNet-101-FPN & 42.8 & 60.3 & 46.1 & 24.1 & 45.7& 54.5 \\
			\bottomrule
	\end{tabular}%
	\label{tab:addlabel}%
\end{table*}%

\subsection{Implementation details}
For fair comparisons, all experiments are implemented based on PyTorch and mmdetection toolbox \cite{mmdetection}. ResNet-50 and ResNet-101 \cite{resnet} are adopted as backbones in our experiments. We use 2 GTX 1080Ti GPUs and 2 images per GPU in all experiments. On mmdetection, the default learning rate is set as 0.02 and 0.01 for MS COCO and PASCAL VOC, respectively, with a batch size of 16 (8 GPUs and 2 images per GPU). As only 2 GPUs are available, we are supposed to divide the learning rate by 4 according to the Linear Scaling Rule \cite{goyal2017accurate}. However, considering the increase in the number of positive samples, we choose to double the learning rate. Thus the learning rate is initialized as 0.01 and 0.005 for MS COCO and PASCAL VOC respectively. We use the SGD as the optimizer for model learning and train all models for 12 epochs. All other hyper-parameters follow the settings in mmdetection if not specifically noted.

As for the hyper-parameters of generating RoI samples, we set $\bm{N=4, M=64}$. It means that we split the IoU range into 4 intervals: [0.5, 0.6), [0.6, 0.7), [0.7, 0.8), [0.8, 1.0) and each GT bounding box will generate 64 samples. For the regression loss weight for different IoU intervals, $\bm{{w_1} = 1.0, {w_2} = 1.5, {w_3} = {w_4} = 3.0}$, We also tried larger N and split the IoU range into more elaborate intervals but did not obtain noticeable improvements.

\subsection{Main results}
{\bf Experiments on PASCAL VOC.} The original evaluation metric of PASCAL VOC is to calculate the mAP at 0.5 IoU threshold. As our methods is mainly designed to alleviate the performance imbalance among different IoU levels, we extend the original metric to the COCO-style criterion which calculates the average AP across IoU thresholds from 0.5 to 0.95 with an interval of 0.05. We evaluate the validity of our proposed method on two state-of-the-art object detectors: Faster R-CNN and Cascade R-CNN. From Table 1, we can see that the performance improvement on Faster R-CNN is obvious, 5.2 and 4.8 points higher with the backbone of 50 and 101 layers respectively. We can also find that even though both of Cascade R-CNN and our method solve the same IoU distribution imbalance problem in different ways, our method still raises their performance. These results further demonstrate the compatibility and adaptivity of our method. Further analysing the performance on different IoU thresholds, we can observe that most of the improvements come from the high IoUs, which is in accordance with our expectation.

{\bf Experiments on MS COCO.} To further demonstrate the generalization capacity of our approach, we also conduct experiments on more challenge COCO dataset. All reported results follow standard COCO-style Average Precision (AP) metrics, AP$_{50}$ (AP for IoU threshold 50$\%$), AP$_{75}$ (AP for IoU threshold 75$\%$). We also include AP$_{S}$, AP$_{M}$, AP$_{L}$, which correspond to the results on small, medium and large scales respectively. Table 2 shows the resuls on validation set, our approach brings consistent and substantially improvement across multiple detectors with different backbone. Specially, it improves Faster R-CNN and Cascade R-CNN by 2.7 and 0.5 points on ResNet-50-FPN backbone and 2.4 and 0.4 points on ResNet-101-FPN backbone.

{\bf Comparison with state-of-the-art methods.} We compare IoU-uniform R-CNN with the state-of-the-art object detection approaches on the COCO test-dev in Table 3. Without bells and whistles, it achieves 41.2 AP with ResNet-101-FPN, which is 2.4 points higher than the baseline. With more powerful feature extractor and base detector, IoU-uniform R-CNN achieves 42.8 AP, demonstrating the superior performance of our method.
\subsection{Analysis}
We perform a thorough study on each component of our method and explain how it works from complete statistics.

\begin{table}[htbp]
	\centering
	\caption{Effects of each component in our IoU-uniform R-CNN (EFO means eliminate feature offsets). Results are reported on PASCAL VOC 2007.}
	\setlength{\tabcolsep}{0.5mm}{
		\begin{tabular}{lcccc}
			\toprule
			method & AP    & AP$_{50}$  & AP$_{75}$  & Improvement \\
			\midrule
			Baseline & 50.2  & 79.1  & 54.7  &  \\
			+uniform IoU distribution & 52.1  & 79.8  & 56.2  & +1.9/+0.7/+1.5 \\
			+EFO(with IoU predictor) & 54.4  & 79.5  & 59.6  & +4.2/+0.4/+4.9 \\
			+tuning weight & 55.4  & 79.8  & 59.6  & +5.2/+0.7/+4.9 \\
			\bottomrule
	\end{tabular}}%
	\label{tab:addlabel}%
\end{table}%

{\bf Component Analysis.} To analyze the importance of each proposed component, we report the overall ablation studies in Table 4. We gradually add our strategies on Faster R-CNN with ResNet-50 FPN and report the results on PASCAL VOC 2007. We can learn that the generated IoU uniformly distributed samples empower the detector to have more potential to make process. Each component of our method obtain gains and the combination of them achieves a total gain of 5.2 AP.

{\bf How dose the generated training samples affect the regressor?}  To verify the improvement it brings to the regressor, we first couducted a check experiment that split the classifier and regressor into two branch and found its performance drops a little, which exclude the factor of spliting the original branch. As shown in Figure 3, compared with the original, our regressor significantly improves its performance in the face of high IoU proposals. Take it a step further, we plot the IoU histogram of bbox after the refinement of regressor in Figure 6. It can be seen that we obtain more high IoU bboxes by using the IoU uniformly distributed samples for training.

\begin{figure}[t]
	\begin{center}
		\includegraphics[width=1.0\linewidth]{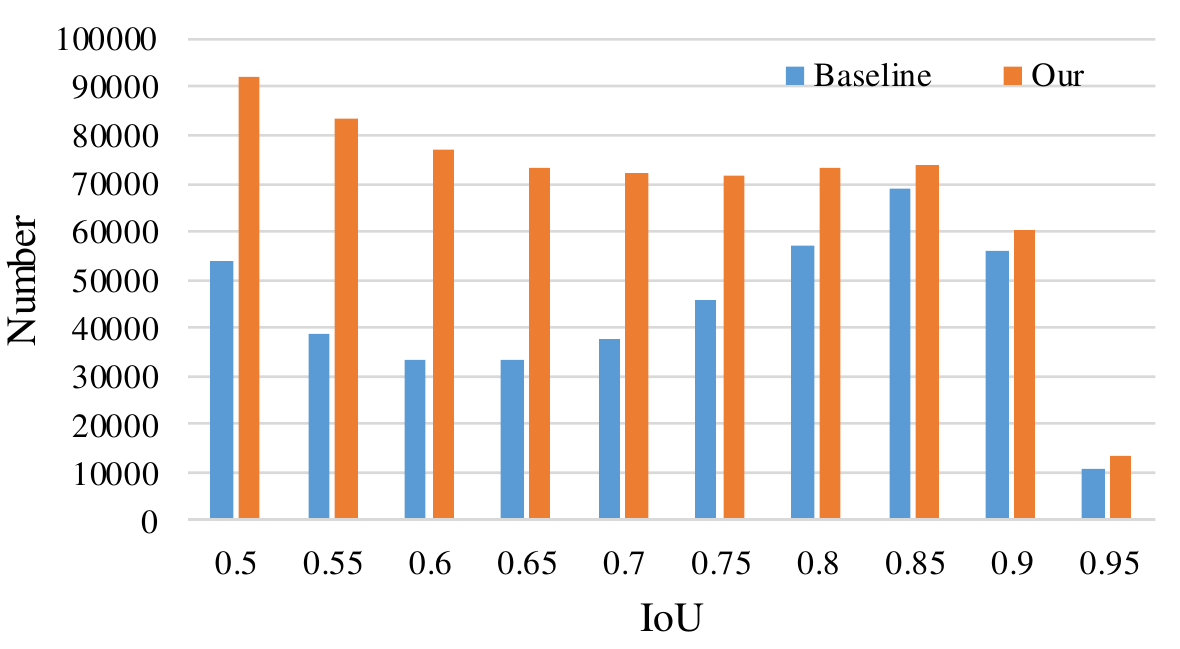}
	\end{center}
	\caption{The IoU histogram of bboxes.}
	\label{fig:long}
	\label{fig:onecol}
\end{figure}

{\bf How dose the generated training samples affect the IoU predictor?} To verify the validity of training IoU predictor with IoU uniformly distributed samples, we only use the generated samples to train the IoU predictor, and the training samples of the regression branch still come from the output of RPN. For comparison, we also train a model that using the output of RPN as the training samples of IoU predictor. It can be seen in Table 6 that the detector whose IoU predictor was trained with the generated sample outperform the one trained with the output of RPN by 2.3 points. With a more powerful IoU predictor, we obtain a more reliable metric to rank the bounding boxes, which would promote the proposal reservation in NMS. To analyse the improvement, we plot the recall curve for different NMS algorithms in Figure 7, with the matching IoU ranging from 0.5 to 1. We can find it achieves better recall among different IoU thresholds, indicating that it helps the NMS process to preserve accurately localized bbox.

{\bf Influence of uniform IoU distribution and increasing number of samples.} Although we have already shown the validity of using generated training samples, we are still unable to determine whether the improvement is mainly from uniform distribution or just the increasing number of samples. Hence we design a experiment that we resample the generated samples. The quantity of generated samples depends on the number of original positive samples produced by RPN. As we can see in Table 5, the IoU uniformly distributed samples can achieve 3.2 improvements of AP with the same number of samples. If we double the quantity, the performance can be further improved with 1.7 points. From above results, we can identify IoU imbalance as the main obstacle to obtain better performance of existing network structure and we can also obtain additional gains from more RoI samples.

{\bf Influence of tuning the regression weight.} As discussed in section 3.1, we can construct a more balanced regression loss among different IoU intervals by tuning the loss weight of different IoU intervals. This is also supported by our experiments. As shown in Figure 3, the performance of regressor has been further improved.

\begin{table}[htbp]
	\centering
	\caption{The results of training with different number of generated samples on PASCAL VOC 2007 dataset}
	\setlength{\tabcolsep}{1.0mm}{
		\begin{tabular}{ccccc}
			\toprule
			Method  & Num of samples  & AP    & AP$_{50}$ & AP$_{75}$  \\
			\midrule
			Baseline          &       & 50.2  & 79.1  & 54.7 \\
			IoU-uniform R-CNN & Equal & 53.4  & 79.3  & 57.5 \\
			IoU-uniform R-CNN & Double& 55.1  & 80.3  & 59.5 \\
			\bottomrule
	\end{tabular}}%
	\label{tab:addlabel}%
\end{table}%

\begin{table}[htbp]
	\centering
	\caption{Studies on the effects of eliminating feature offsets and training IoU predictor with generated samples on PASCAL VOC.}
	\setlength{\tabcolsep}{0.5mm}{
		\begin{tabular}{ccc}
			\toprule
			Training samples & Eliminating feature offsets & AP \\
			\midrule
			\multirow{2}[2]{*}{Output of RPN} & No    & 48.7 \\
			& Yes   & 49.9 \\
			\midrule
			\multirow{2}[2]{*}{Generated samples} & No    & 50.2 \\
			& Yes   & 52.2 \\
			\bottomrule
	\end{tabular}}%
	\label{tab:addlabel}%
\end{table}%

{\bf Influence of eliminating the feature offsets.} From the results reported in Table 6, we can find its major impact on the final performance, as the gain from IoU uniformly distributed samples is almost offset without eliminating the feature offsets. Figure 8 may answer the question why it has such huge impact on the performance. The x-axis is the IoU between the refined bbox and its matched ground-truth, while the y-axis denotes its predicted value. We can find in Figure 8 (a) that the predicted value is not well correlated with the ground truth. We attribute this to the transition of low IoU to high IoU. As the average IoU increment for those low IoU-level candidate bbox is around 0.2, the unupdated predicted value is far behind the ground truth. This leads to the potential suppression of accurate located bbox. Visualized in Figure 8 (b), the IoU estimation becomes more accurate after eliminating the feature offset. It is worth mentioning that the stronger the IoU prediction branch is, the more gains it can bring. The gain of generated uniform samples is 2.0 points compared to 1.2 of RPN samples. The qualitative results for comparison between the eliminating feature offsets with unequipped one are provided in Figure 9. We can see that eliminating the feature offsets of RoIs can help preserve more accurate detection results. These further demonstrate the effectiveness of eliminating the feature offsets of RoIs for better IoU prediction.

\begin{figure}[t]
	\begin{center}
		\includegraphics[width=0.8\linewidth]{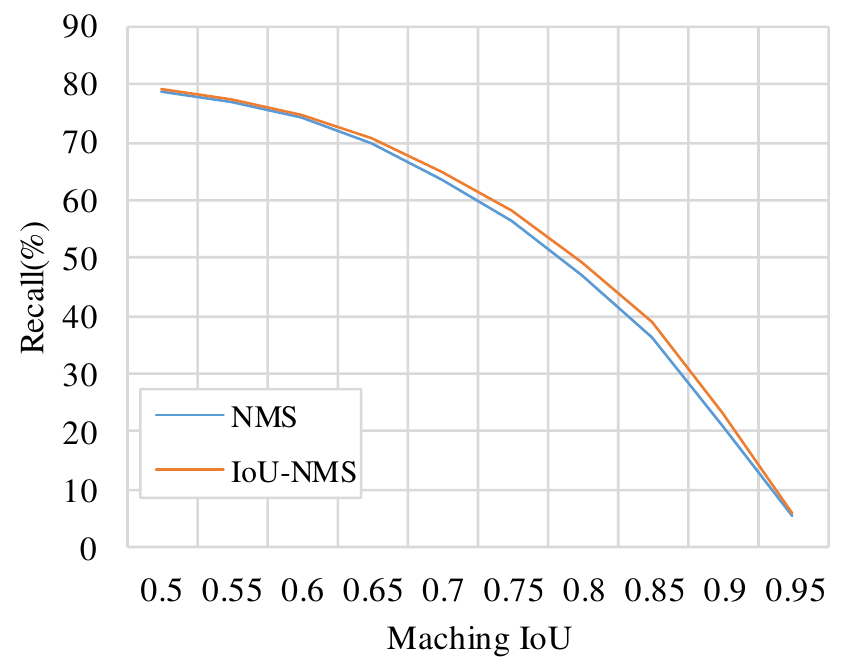}
	\end{center}
	\caption{Comparison among the recall curves of different NMS
		algorithms with the matching IoU ranging from 0.5 to 1.}
	\label{fig:long}
	\label{fig:onecol}
\end{figure}

\begin{figure}[t]
	\begin{center}
		\includegraphics[width=0.8\linewidth]{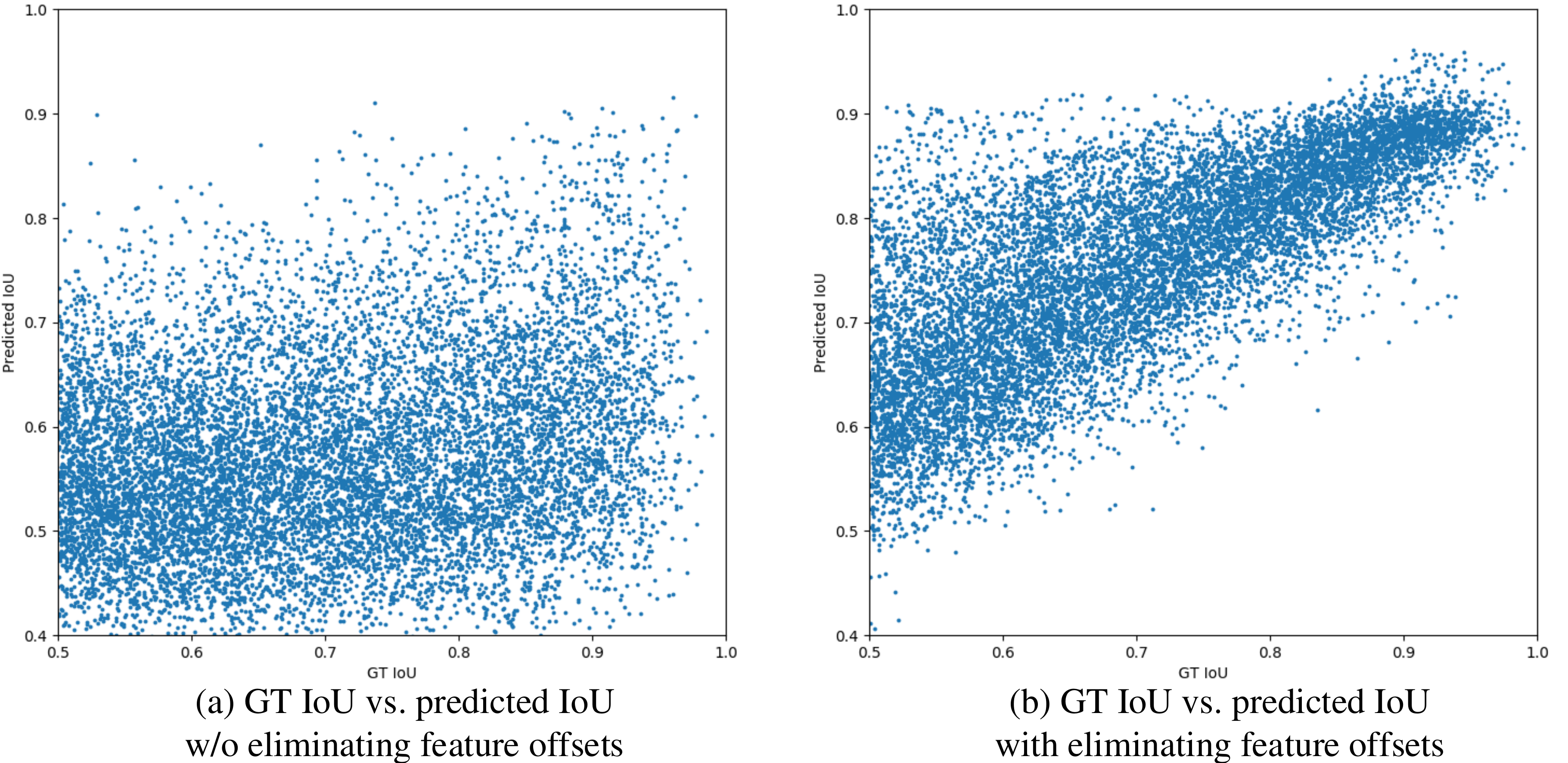}
	\end{center}
	\caption{The correlation between the IoU of bounding boxes with the corresponding ground-truth and the predicted IoU. Considering detected bounding boxes having an IoU (\textgreater 0.5) with the corresponding ground-truth.}
	\label{fig:long}
	\label{fig:onecol}
\end{figure}

\begin{table}[htbp]
	\centering
	\caption{Ablation analysis on the number of images per GPU and the learing rate on PASCAL VOC 2007 dataset}
	\setlength{\tabcolsep}{0.5mm}{
		\begin{tabular}{cccccccc}
			\toprule
			Num  & Lr    & AP    & AP$_{50}$ & AP$_{60}$  & AP$_{70}$  & AP$_{80}$  & AP$_{90}$ \\
			\midrule
			2     & 0.0025 & 53.18 & 79.3 & 73.7 & 64.5 & 48.9 & 19.9  \\
			$\bm{2}$ &  $\bm{0.005}$ &  $\bm{54.42}$ & 79.5  & 74.4  & 66.0    & 50.1  & 21.8 \\
			4     & 0.005 & 53.44 & 79.0    & 74.1  & 64.8  & 48.3  & 18.2 \\
			4     & 0.01  & 54.19 & 79.7  & 74.7  & 65.6  & 50.5  & 21.5 \\
			4     & 0.015 & 53.84 & 78.6  & 73.9  & 65.5  & 49.7  & 21.4 \\
			\bottomrule
	\end{tabular}}%
	\label{tab:addlabel}%
\end{table}%

\begin{figure}[t]
	\begin{center}
		\includegraphics[width=0.9\linewidth]{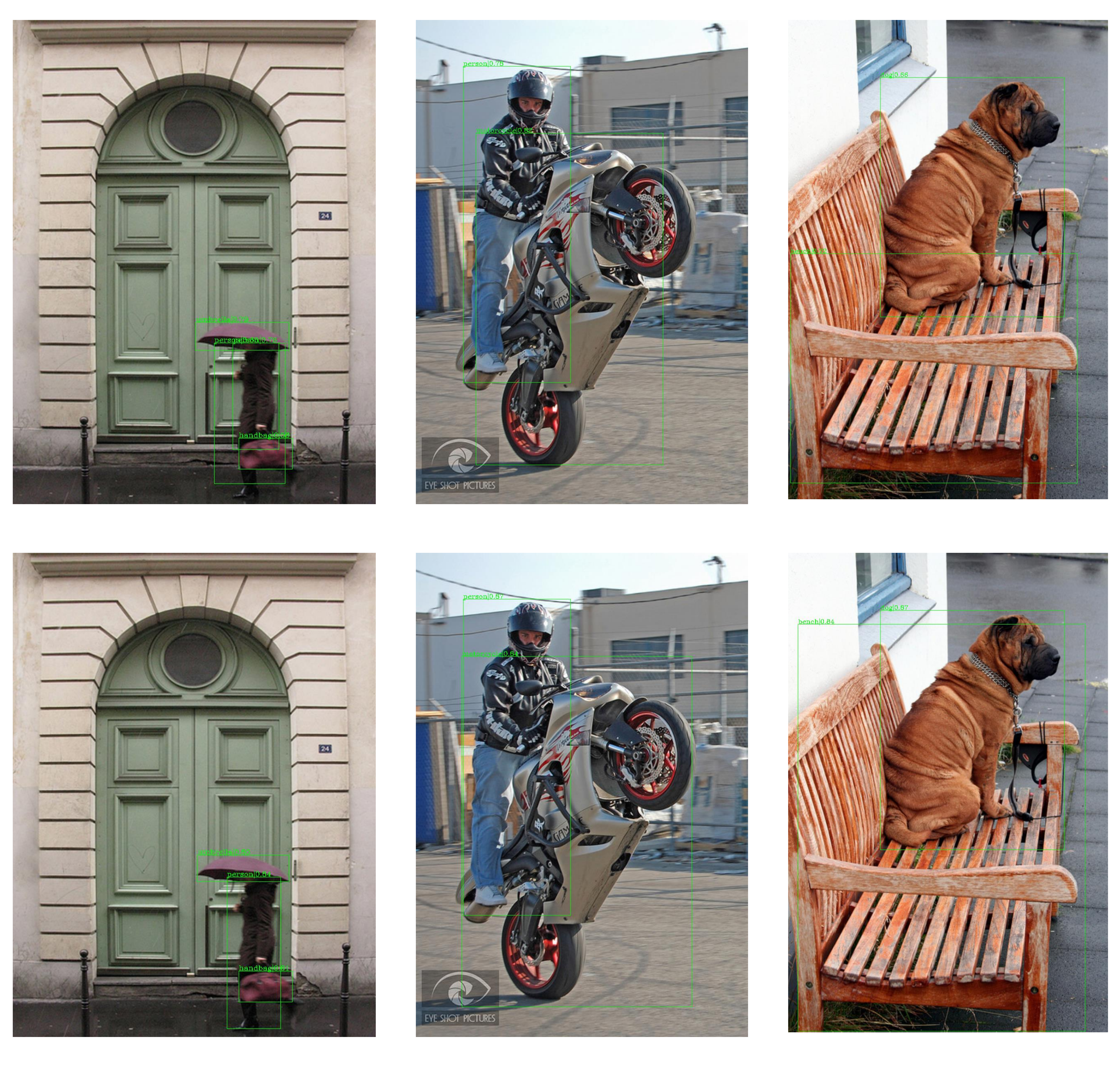}
	\end{center}
	\caption{Qualitative comparison between the eliminating feature offsets with unequipped one on MS COCO dataset. The first row shows the results without eliminating feature offsets and use the first time value as the final IoU prediction. The second row shows the results with eliminating feature offsets.}
	\label{fig:long}
	\label{fig:onecol}
\end{figure}

\subsection{Ablation study} 
During training, we found that the performance of our IoU-uniform R-CNN is sensitive to the batch size and learning rate. According to Linear Scaling Rule, the learning rate is supposed to be divided by 4, as we only have 2 GPUs available compared with the default 8 GPUs. Thus, for Faster R-CNN with ResNet-50-FPN backbone on PASCAL VOC, the learning rate is supposed to be 0.0025. But we found that we can obtain better results by double the learning rate. We attribute this to the increasing training samples for regression branch. The number of original training samples for regression branch is usually no more than 100 per image, but when we generate samples by ourselves, the average number of samples reaches 200. We further conduct ablation studies on the number of images per GPU and learning rate to determine the suitable hype-parameters. It is shown in Table 7 that we obtain the best results by setting the num=2 and learning rate=0.005.

\section{Conclusion}
In this paper, we reveal the limitations of RPN and rethink the IoU distribution imbalance problem in object detection. The proposed IoU-uniform R-CNN, a simple but effective way, alleviates the imbalance in both the number of samples and regression loss among different IoU intervals. In particular, we first replace the RoI training samples with generated IoU uniformly distributed samples. Then we tune the loss weight of different IoU intervals to further control the balance of regression loss composition. Besides, we also point out the feature offsets of RoIs during the inference of IoU-prediction branch and solve it by updating the feature of refined RoIs. Extensive experiments show its superior performance on both PASCAL VOC and MS COCO dataset, as well as its compatibility and adaptivity to many object detection architectures.

{\small
\bibliographystyle{ieee_fullname}
\bibliography{egbib}
}

\end{document}